\documentclass[10pt,twocolumn,letterpaper]{article}

\usepackage{iccv}
\usepackage{times}
\usepackage{epsfig}
\usepackage{graphicx}
\usepackage{multirow}

\usepackage{float}
\usepackage{notoccite}
\usepackage[numbers,sort&compress]{natbib}
\usepackage{amsmath}
\usepackage{amssymb}
\usepackage[table,xcdraw]{xcolor}
\usepackage{subcaption}
\usepackage[group-separator={,}]{siunitx} 

\usepackage[breaklinks=true,bookmarks=false]{hyperref}

\iccvfinalcopy 

\def\iccvPaperID{****} 

\ificcvfinal\fi

\begin{document}

\title{Guarding the Guardians: Automated Analysis of Online Child Sexual Abuse}

\author{
Juanita Puentes$^{1}$ \quad
Angela Castillo$^{1}$ \quad
Wilmar Osejo$^{2}$ \quad
Yuly Calderón$^{3}$ \\
Viviana Quintero$^{2}$ \quad
Lina Saldarriaga$^{2}$ \quad
Diana Agudelo$^{3}$ \quad
Pablo Arbel\'aez$^{1}$ \\
$^{1}$Center for Research and Formation in Artificial Intelligence, Universidad de los Andes \\ $^{2}$Aulas en Paz \quad $^{3}$Universidad de los Andes 
}

\maketitle
\ificcvfinal\thispagestyle{empty}\fi

\begin{abstract}

Online violence against children has increased globally recently, demanding urgent attention. Competent authorities manually analyze abuse complaints to comprehend crime dynamics and identify patterns. However, the manual analysis of these complaints presents a challenge because it exposes analysts to harmful content during the review process. 
Given these challenges, we present a novel solution, an automated tool designed to analyze children's sexual abuse reports comprehensively. By automating the analysis process, our tool significantly reduces the risk of exposure to harmful content by categorizing the reports on three dimensions: Subject, Degree of Criminality, and Damage. Furthermore, leveraging our multidisciplinary team's expertise, we introduce a novel approach to annotate the collected data, enabling a more in-depth analysis of the reports. This approach improves the comprehension of fundamental patterns and trends, enabling law enforcement agencies and policymakers to create focused strategies in the fight against children's violence.

\end{abstract}
\section{Introduction}
Since 2020, WeProtect~\cite{Protect_CRISP_PA} has reported a significant increase in the production of Child Sexual Exploitation and Abuse (CSEA) worldwide. 
Alarming rises have been found in the frequency of grooming cases, the volume of CSEA, the distribution of material on the dark web, and the live streaming of abuse. 
In 2021, the National Center for Missing and Exploited Children (NCMEC)~\cite{NCMEC} processed around \num{44155} online enticement cases, increasing to \num{80524} in 2022.
Similarly, between 2021 and 2022, Te Protejo, a Colombian Hotline, processed nearly \num{1196} reports on sextortion, sexting, grooming, and sexual cyberbullying. 
Despite numerous studies analyzing the production of Child Sexual Abuse Material (CSAM) worldwide, there is limited evidence regarding its characteristics and dynamics in Latin America. 
Moreover, effective practices to aid authorities and hotlines in analyzing information to combat this crime in Spanish-speaking countries remain scarce.

Manually processing reports received by authorities or hotlines is a significant bottleneck in identifying crimes. Challenges include data variability, extensive information in aggressive conversations, and classifying cases into crime categories. It takes 25 minutes for an analyst to identify the Subject, Degree of Criminality, and Damage in a single report. Thus, the intensive exposure of analysts to harmful content is a significant concern as it can adversely impact their mental health and well-being~\cite{Ahern2017WoPW}. Please refer to the Supplementary Section~\ref{cybercrime} for further details. 

We propose a Large Language Model (LLM) that analyzes the information received in reports on sextortion, sexting, grooming, and sexual cyberbullying by Te Protejo. Our system analyzes, classifies, and efficiently forwards the reports to competent authorities. By reducing analysts' exposure to harmful content, our solution enhances their mental well-being, ensuring a safer working environment.


	




\section{Related Work}
\subsection{Online Violence Against Children}
Recently, there has been an increase in the production of studies that analyze CSAM. 
In the last four years, at least 43 articles have been published detailing the sources, methods, and types of information about the production of CSAM~\cite{Ngo}. 
However, none of these refer to work done specifically for the Latin American context.

Several initiatives aim to assist hotlines in efficient data collection and report processing while reducing the harm caused by analysts' exposure to information.  
One of these initiatives is the Aviator Project (Augmented Visual Intelligence and Targeted Online Research), a tool that helps hotlines and law enforcement agencies process, assess, and prioritize CSAM reports.
This work, supported by 16 police agencies from all over Europe, such as Ziuz Forensic, Web IQ, INHOPE, and the European Union Police Internal Security Fund, has shown promising advances in using visual intelligence to prioritize reports and conduct analysis more efficiently.  
However, up to date, there are no specific tools designed to help hotlines and analysts in Latin America\footnote{In the Latin American context, there are four reporting lines: Te Protejo Colombia, Te Protejo México, SaferNet in Brazil, and Grooming Argentina. In addition, there are web portals in El Salvador, Guatemala, Argentina, the Dominican Republic, Belize, and Haiti run by the Internet Watch Foundation (IWF).} to process the information. 
Added to the limited understanding of the phenomenon in Spanish, designing better reporting and analysis mechanisms, such as those that Artificial Intelligence models can provide, is necessary.

\begin{figure}[t]
    \centering
    \includegraphics[scale = 0.17]{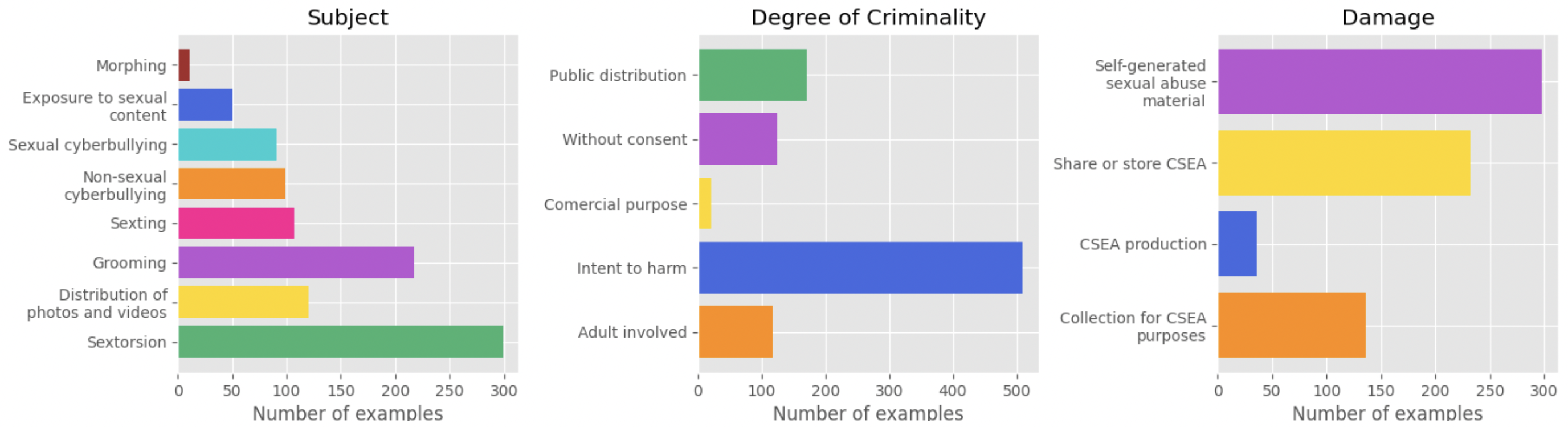}
    \caption{\textbf{Distribution of Classes in the Subject, Degree of Criminality, and Damage Data:} 
    Our dataset exhibits significant class imbalance, with certain classes containing more samples than others.
    }
    \label{fig:stats}
\end{figure}

\begin{figure*}[t]
\centering
\begin{center}
\end{center}
    \centering
    \includegraphics[width=0.85\textwidth]{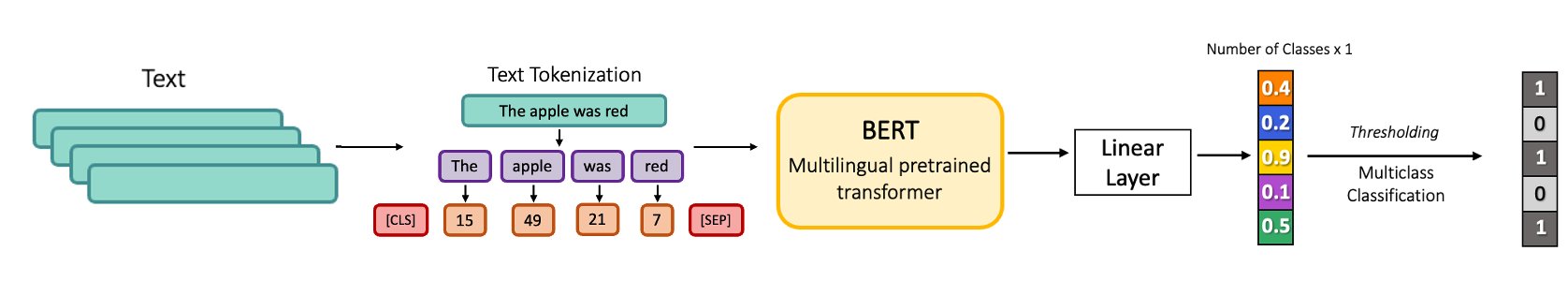}
   \caption{\textbf{Overview of the LLM method.} Our method uses a BERT-based~\cite{devlin2019bert} model to produce a prediction based on our pre-processed complaints. Then, we provide a multiclass prediction according to the categories in each dimension. }
\label{method}
\end{figure*}

\subsection{Large Language Models} 
Since introducing Bidirectional Encoder Representations from Transformers (BERT)~\cite{devlin2019bert} in 2018, language models showed better transfer learning capabilities, more contextualized word embeddings, and multilingual support.
Concurrently, the Generative Pre-trained Transformer (GPT)~\cite{yenduri2023generative} family introduced the concept of using a transformer-based architecture for generating text.  
One year later, XLNet~\cite{yang2020xlnet} considered all possible permutations of the input sequence during training, while RoBERTa~\cite{liu2019roberta} and ALBERT~\cite{lan2020albert} proposed better more efficient models.
Recent work, such as GPT-3 and CLIP~\cite{radford2021learning}, focused on Large Language Model (LLM) training that uses an advanced language model at an impressive scale.
However, despite the progress in language processing techniques, there is a notable gap in specializing these powerful methods for online child abuse detection.
\section{Experimental Setup}
\subsection{Dataset}
We gather complaints from a CSEA Hotline, a reporting line dedicated to safeguarding the rights of children and adolescents in Colombia. We filter the complaints about children's online abuse from victims, parents, and other stakeholders. With these complaints, we create a dataset from the Hotline from January 2021 to December 2022. Specifically, we use \num{1196} reports from the Sexual Abuse Material category related to grooming, sextortion, disclosure of sexual content, and cyberbullying.

\par{\textbf{Expert Annotations.}}
We employ comprehensive data analysis to establish a robust annotation system that trains our prediction models effectively.
For this purpose, we have engaged experts with a wealth of experience and expertise, qualifying them as ideal candidates for annotating complaints. Moreover, the experts produce an in-depth analysis to categorize the complaints in three dimensions following the guidelines from We Protect~\cite{Protect_CRISP_PA}.
A complaint is associated with multiple labels within each dimension (Subject, Degree of Criminality, and Damage), allowing for a multiclass approach.

Figure~\ref{fig:stats} includes the final classification categories across the dimensions. The Subject dimension comprises a total of \num{994} data instances distributed across \num{8} classes. Among these, the \textit{sextortion} class exhibits the highest data instances, totaling \num{299}. In the Degree of Criminality, there are \num{943} data instances, with over half belonging to the \textit{intent of damage} class. The least represented class in this dimension is \textit{commercial purpose}, with only \num{21} data instances. Lastly, the Damage consists of a total of \num{702} data instances distributed across \num{4} classes.

The interaction between the Machine Learning (ML) researchers and the experts is crucial because the ML researchers do not possess direct access to the raw data, given its sensitivity. Therefore, the validation of the method is performed in two stages. Firstly, on the part of the ML researchers responsible for developing the LLM method, the model's input is considered a singular row originating from a spreadsheet file. The ML researchers produce an initial prediction based on the raw input.
Then, once they have the prediction, the experts in the area are in charge of analyzing the failing cases of the NLP model since they are exclusively granted access to read the complaints and conduct an exhaustive analysis of each case. The experts provide insights into the failure cases, and the ML researchers develop a strategy to improve the model. This process is done iteratively to improve the predictive model. Consequently, we highlight the significance of a multi-disciplinary team.

\par{\textbf{Data pre-processing.}}
The dataset comprises complaints encompassing personal identifiers such as telephone numbers, ID numbers, emails, and URLs. Therefore, to ensure data anonymity and privacy, we systematically eliminate these sensitive details from the dataset.

\par{\textbf{Evaluation metrics.}}
As our data is highly imbalanced, we perform multiclass binary classification. Thus, we implement the Precision-Recall (PR) curves to fully exploit the benefits of modeling the classification task as a detection to assess the performance of our classification problem. Furthermore, we employ the traditional F-score as the evaluation metric, which measures the harmonic mean of the precision and recall. Our experiments evaluate the F-score and mean Average Precision (mAP), calculated as the average area under the curve for all the classes.

Our model categorizes the complaints based on the different classes within each dimension. We proceed with our data using a two-fold cross-validation strategy. We randomly shuffle the dataset into two sets, ensuring that both sets are of equal size and maintain the class proportion from the original dataset in both folds. We report the average and standard deviation between our two sets.

\subsection{Model}
We extend the traditional classification transformer architecture by adapting the classification head according to each dimension, as shown in Fig.~\ref{method}. Therefore, since we aim to fully exploit the information of each complaint to study them on each dimension, we train individual models for the three dimensions. The input to our models consists of tokenized complaints using the BERT Multilingual Tokenizer~\cite{devlin2019bert}. Then, we utilize a linear classification layer to assign a probability to each class. By framing this classification task as a detection problem, we use various thresholds to assign positive labels based on the output probabilities to optimize a binary cross-entropy loss.
Our optimization protocol enforces the non-exclusive category prediction, which aligns well with the nature of our problem.

\subsection{Learning}
Given the sensitive nature of complaints of sexual abuse on children, we encounter restrictions in the amount of available data for its analysis. 
Thus, we employ data augmentation to alleviate the limited data of the available instances. We create augmented training datasets of different sizes by applying a data augmentation technique, specifically deleting random words. Each augmented complaint has words removed with varying probabilities between \num{0.05} to \num{0.9}, enabling a more comprehensive range of experimentation and mitigating the lack of data. Thus, we optimize the Augmentation Deletion Rate (ADR) parameter, which is the likelihood of word deletion during augmentation.
\section{Results}

\par{\textbf{Implementation Details.}}
We use a BERT multilingual model~\cite{devlin2019bert}, a pretrained model on the top \num{104} languages with the largest Wikipedia. 
In that way, we could use this pre-trained model for this Spanish NLP classification task.   

\subsection{Baseline}

\begin{table*}[h]
\centering
\setlength{\tabcolsep}{14pt}
\resizebox{17.3cm}{!}{\begin{tabular}{cccccccc}
\hline
                &       & \multicolumn{2}{c}{\textbf{Baseline}} & \multicolumn{2}{c}{\textbf{Fine-Tuning}} & \multicolumn{2}{c}{\textbf{Data Augmentation}} \\ \hline
\rowcolor[HTML]{EFEFEF} 
\textbf{Dimension}   &\textbf{Parameters}       & mAP        & F-score       & mAP         & F-score         & mAP            & F-score            \\ \hline
Subject             &177M  & $0.148 \pm 0.005$           &  $0.192    \pm 0.035$              & $0.382 \pm 0.016$           & $\mathbf{0.455 \pm 0.001}$                & $\mathbf{0.386 \pm 0.002}$               &  $0.447 \pm 0.014$                   \\
\rowcolor[HTML]{EFEFEF} 
Degree of Criminality&177M & $ 0.296   \pm 0.026$           & $0.600     \pm 0.009$             & $0.397 \pm 0.040$            & $0.593 \pm 0.017$                & $\mathbf{0.417 \pm 0.032}$            & $\mathbf{0.598 \pm 0.036}$                    \\
Damage                &177M& $0.338  \pm  0.027$        & $0.525     \pm  0.076$          & $0.429 \pm 0.018$          & $0.553 \pm 0.004$                & $\mathbf{0.456 \pm 0.001}$             & $\mathbf{0.576 \pm 0.019}$                 \\ \hline
\end{tabular}}
\caption{\textbf{Baseline, Fine-Tuning, and Data Augmentation:} We present the mAP and F-score results for the Subject, Degree of Criminality, and Damage dimensions. Note the improvement as we specialize the model to our data corpus domain.}
\label{tab:final}
\end{table*}

\begin{figure}[t]
    \centering
    \includegraphics[scale = 0.17]{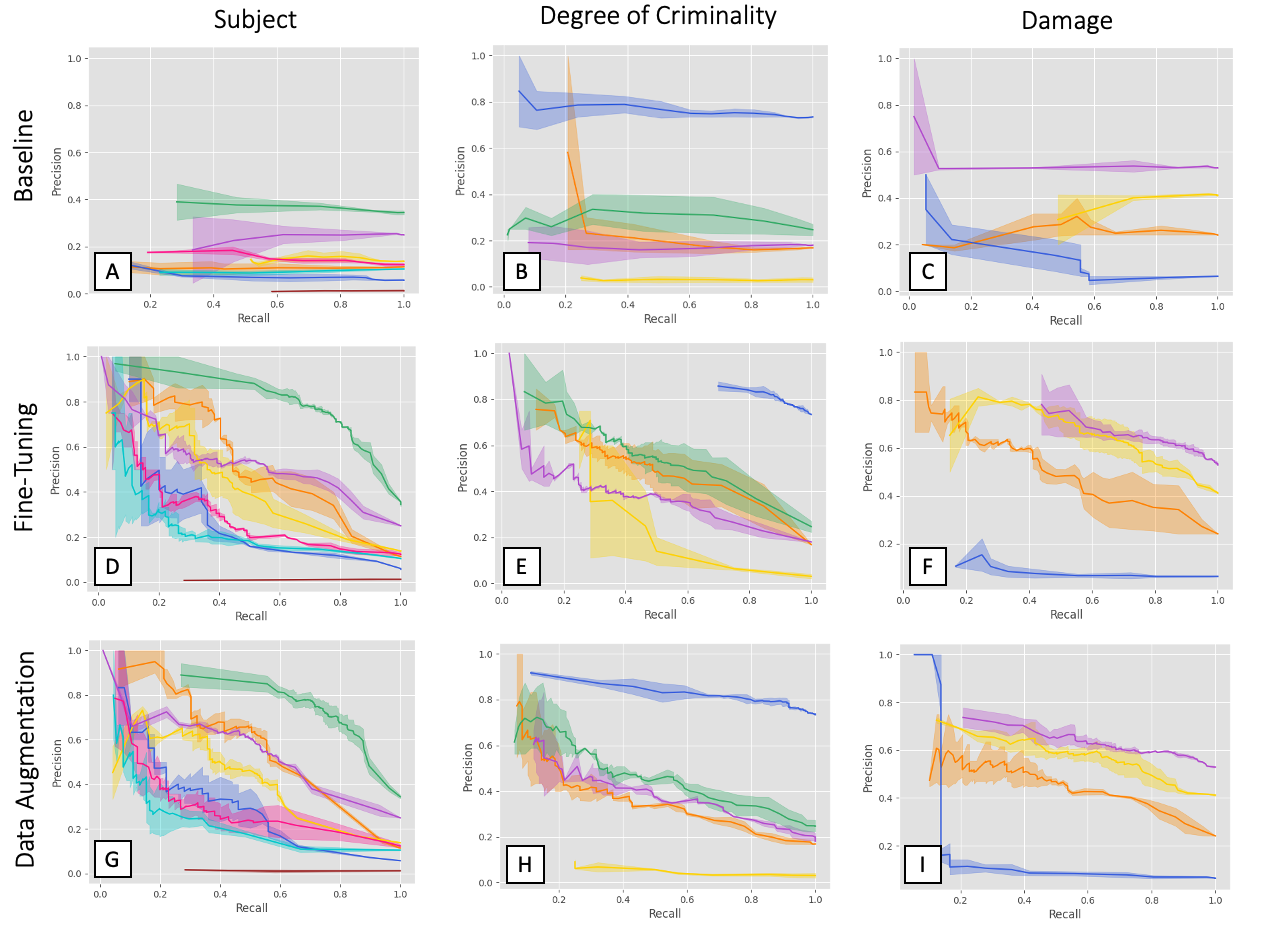}
    \caption{\textbf{Baseline, Fine-Tuning and Data Augmentation:} We present the precision-recall curves for Subject, Degree of Criminality, and Damage dimensions. Observe significant performance improvements with fine-tuning and data augmentation compared to the baseline. The colors assigned to each curve coincide with those employed in Fig.~\ref{fig:stats} for individual classes.}
    \label{fig:curves}
\end{figure}

Table~\ref{tab:final} and Fig.~\ref{fig:curves}A-C present the baseline performance of the pretrained model without additional training. 
The curves for the baseline in Fig.~\ref{fig:curves}A-C for the individual classes show that most of them achieve a consistent precision rate, regardless of the recall attained. 
The presence of a horizontal line in the PR curve, as depicted in most instances in Fig.~\ref{fig:curves}A-C, indicates that the fraction of false positives is constant, which means that the model has reached its performance limit. 
This result suggests that the model's discriminatory ability between positive and negative instances is restricted, resulting in a stable precision rate. Although the baseline approach is an appropriate starting point for addressing this problem, its efficacy is limited for this particular task. Thus, it is imperative to perform fine-tuning using our specific dataset during  training.

\subsection{Model Specialization}
We conduct a subsequent experiment of specializing the model by fine-tuning it on our dataset. For each dimension (Subject, Degree of Criminality, and Damage), we determine the optimal combination of hyperparameters that maximize the mAP. Moreover, we preprocess the complaints to remove unnecessary information such as emails, URLs, and IDs. Upon comparing the fine-tuning results from Table~\ref{tab:final} with the baseline outcomes, a notable improvement is evident, particularly in the Subject dimension. The model demonstrates enhancements in terms of mAP for all the dimensions. Figure~\ref{fig:curves}D-F highlights the effectiveness of fine-tuning in enhancing the model's classification and identification capabilities for specific classes in all dimensions. This fine-tuning also reduces false positive rates, enabling better differentiation between classes and improved identification of relevant examples. Furthermore, our method demonstrates robustness to false negatives, as the recall remains unaffected since the model achieves higher precision without sacrificing recall, and vice versa.

Despite the improvement, the presence of class imbalance becomes apparent. For instance, in the Subject dimension, the PR curves for the \textit{Morphing} class (Fig.~\ref{fig:curves}D), which comprises only \num{11} complaints, do not exhibit improvement compared to the baseline. In the case of the Criminality Degree dimension, the precision curve for the \textit{Commercial purpose} class (Fig.~\ref{fig:curves}E) shows significant enhancement compared to the baseline, despite containing only \num{21} complaints. Nevertheless, it is crucial to consider the deviation observed across different folds for this class.

\subsection{Data Augmentation}
From PR curves (Fig. ~\ref{fig:curves}G-I), we notice a decrease in fold-wise deviation compared to previous experiments. This decrease can be attributed to data augmentation, which introduces heightened diversity and variability into the training data. We generate multiple augmented versions of each original complaint by randomly deleting words in the complaints. However, it is important to acknowledge that the limited improvement in performance indicates the need to consider the potential manifestation of overfitting. Similar to the findings observed during fine-tuning, data augmentation implies a reduction in false positives.

Considering the stochastic nature of the data augmentation procedure, which involves both random complaint selection and the deletion of random words, an inherent probabilistic bias emerges. Classes with a larger initial data volume tend to retain a greater proportion of instances after augmentation. As a result, this disparity manifests as a decline in performance for classes characterized by limited data samples, specifically the \textit{Morphing} class in the Subject dimension (Fig.~\ref{fig:curves}G) and the \textit{Commercial purpose} class in the Damage dimension (Fig.~\ref{fig:curves}H). Conversely, an inverse behavior is observed in the Damage dimension (Fig.~\ref{fig:curves}I), where the augmentation of data leads to improved performance for the \textit{CSEA Production} class.

\section{Conclusion}
This study presents a comprehensive approach encompassing a complete experimental setup and an automated tool to process complaints from online child abuse reports. In particular, we demonstrate an effective way to annotate the complaints to further comprehensively analyze them. Moreover, we show that using a specialized LLM as a predicting tool to categorize the complaints among different dimensions, thus, facilitating a deeper understanding of the dynamics of abuse. Our results show that specializing in a model with a fine-tuning procedure outperforms a traditional LLM. However, it also highlights the need for data augmentation due to the under-representation of this problem. We hope this work will enable the development of new research directed toward this critical problem to advance the knowledge of online violence against children.

\section{Ethical Considerations}
The sensitive nature of the data poses the main ethical consideration: the inability to make our data and methods publicly available. This limitation arises from upholding privacy and confidentiality standards to safeguard the individuals involved. However, we will share our knowledge, methodologies, and models with our partners and other hotlines to improve intervention strategies and advance our understanding of online violence against children.

\subsection*{Acknowledgements}
This project was partially supported by the End Violence Against Children Grant. Data collection for this study was done in partnership with the Te Protejo Reporting Line managed by Red PaPaz.

{\small
\bibliography{egbib}
\bibliographystyle{unsrt}
}

\clearpage

\appendix

\section*{Supplementary Material}

\section{Manual Cybercrime Analysis} \label{cybercrime}
According to the literature, cybercrime analysts are frequently exposed to traumatic experiences and narratives. Exposure to other people's traumatic experiences can lead to vicarious trauma, secondary traumatic stress, compassion fatigue, burnout, and even symptoms of Post Traumatic Stress Disorder~\cite{Molnar2020}. A U.S. federal law enforcement agency study found that 36\% of online CSAM researchers exhibited moderate and high levels of secondary traumatic stress~\cite{Molnar2020}. 

Thus, a manual review of child sexual exploitation and abuse material can affect analysts' physical and mental health, job performance, interpersonal relationships, health, and overall well-being~\cite{Mathieu2012}. Over time, these levels of impairment can impact the accuracy of analysis and, therefore, the prioritization and referral of risk situations to authorities.

Online CSEA cases, on the other hand, have maintained a steady increase over the past few years. In 2020, NCMEC reported a 97.5\% increase in ``online enticement," a form of online exploitation that includes online grooming. Likewise, an analysis of dark web conversations conducted by CRISP found that the number of conversations between sex offenders to share online grooming strategies increased by 13\% between 2019 and 2020. In the same period, NCMEC~\cite{Protect_CRISP_PA} reported a 63\% increase in CSAM reports.

With a problem growing exponentially globally and taking into account the impact on the physical and mental health and well-being of analysts, it is necessary to explore new possibilities to reduce the burden of analysis and increase the processing capacity of reported cases.

\section{Additional Experimental Validation}
In this section, we present a comprehensive overview of the experimental details conducted to assess the performance of our language model. To achieve this, we conduct two sets of experiments: the first set focuses on evaluating the impact of hyperparameters without data augmentation. In contrast, the second set delves into the effects of data augmentation in conjunction with hyperparameters. Both sets of experiments employ fine-tuning.

In the first set of experiments, we thoroughly investigate the influence of critical hyperparameters on the performance of our fine-tuned language model without employing any data augmentation. The hyperparameters examined encompass the learning rate, batch size, dropout rate, and the number of training epochs. The best hyperparameter configurations for each dimension: Subject, Degree of Criminality, and Damage, are summarized in Table \ref{tab:finetuning-hyper}.

\begin{table}[h]
\centering
\setlength{\tabcolsep}{15pt}
\resizebox{8.3cm}{!}{
\begin{tabular}{cccc}
\hline
\textbf{Hyperparameters}   & \textbf{Subject} & \textbf{Degree of Criminality} & \textbf{Damage} \\ \hline
\rowcolor[HTML]{EFEFEF} 
Batch Size Train           & 41              & 167                            & 54             \\
Batch Size Test            & 68              & 39                             & 171             \\
\rowcolor[HTML]{EFEFEF} 
Learning Rate              & 1.217 E-5         & 4.634 E-5                       & 5.804 E-5         \\
Epochs                     & 144              & 116                             & 10             \\
\rowcolor[HTML]{EFEFEF} 
Dropout                    & 0.448           & 0.218                          & 0.485          \\

\textbf{mAP}               & $0.382 \pm 0.0016$         & $0.397 \pm 0.040$                               & $0.429 \pm 0.018$               \\
\rowcolor[HTML]{EFEFEF} 
\textbf{F-score}           & $0.455 \pm 0.001$                & $0.593 \pm 0.017$                              & $0.553\pm 0.004$             \\ \hline
\end{tabular}}
\caption{\textbf{Fine-Tuning experimentation details.} We present the hyperparameters used for Subject, Degree of Criminality and Damage dimensions. }
\label{tab:finetuning-hyper}
\end{table}

In the second set of experiments, we present the impact of data augmentation on the performance of our fine-tuned language model. For this analysis, we carefully identify the optimal hyperparameters specific to the data augmentation configuration. Additionally, we introduced two additional augmentation-specific parameters: the Augmentation Factor (AF) and the Augmentation Deletion Rate (ADR). AF is the multiplier factor used to increase the training dataset size through data augmentation. ADR represents the likelihood of each word being deleted during augmentation. These identified optimal hyperparameter configurations, tailored for data augmentation for each dimension, are summarized in Table \ref{tab:augmentation-hyper}.

\begin{table}[h]
\centering
\setlength{\tabcolsep}{15pt}
\resizebox{8.3cm}{!}{
\begin{tabular}{cccc}
\hline
\textbf{Hyperparameters}   & \textbf{Subject} & \textbf{Degree of Criminality} & \textbf{Damage} \\ \hline
\rowcolor[HTML]{EFEFEF} 
Batch Size Train           & 75               & 221                            & 200             \\
Batch Size Test            & 212              & 89                             & 169             \\
\rowcolor[HTML]{EFEFEF} 
Learning Rate              & 3.569 E-6         & 8.399 E-6                       & 1.212 E-5         \\
Epochs                     & 140              & 13                             & 91              \\
\rowcolor[HTML]{EFEFEF} 
Dropout                    & 0.247            & 0.435                          & 0.498           \\
ADR & 0.098            & 0.061                          & 0.856           \\
\rowcolor[HTML]{EFEFEF} 
AF        & 4.354            & 8.77                           & 1.532           \\
\textbf{mAP}               & $0.386 \pm 0.002$         & $0.417 \pm 0.032$                               & $0.458 \pm 0.001$               \\
\rowcolor[HTML]{EFEFEF} 
\textbf{F-score}           & $0.447 \pm 0.014$                & $0.598 \pm 0.036$                              & $0.576 \pm 0.019$             \\ \hline
\end{tabular}}
\caption{\textbf{Data Augmentation experimentation details.} We present the hyperparameters used for Subject, Degree of Criminality and Damage dimensions. }
\label{tab:augmentation-hyper}
\end{table}

\end{document}